\documentclass{article}
\usepackage{iclr2027_conference,times}

\usepackage{amsmath,amsfonts,bm}

\def\eqref#1{equation~\ref{#1}}
\def\Eqref#1{Equation~\ref{#1}}

\def\1{\bm{1}}

\DeclareMathAlphabet{\mathsfit}{\encodingdefault}{\sfdefault}{m}{sl}
\SetMathAlphabet{\mathsfit}{bold}{\encodingdefault}{\sfdefault}{bx}{n}

\usepackage{hyperref}
\usepackage{url}
\usepackage{booktabs}
\usepackage{float}
\usepackage{placeins}
\usepackage{iclr2027_conference,times}
\usepackage{graphicx}
\usepackage{xcolor}
\definecolor{citationblue}{HTML}{1E00FE}
\hypersetup{
    colorlinks=true,
    citecolor=citationblue,
    linkcolor=citationblue,
    urlcolor=citationblue
}
\definecolor{revisionblue}{HTML}{000000}
\newcommand{\revision}[1]{#1}
\usepackage{multirow}
\usepackage{graphicx,booktabs,amsmath}
\usepackage[T1]{fontenc}

\usepackage{etoolbox}
\makeatletter
\patchcmd{\@maketitle}{{\LARGE\sc \@title\par}}
  {\toptitlebar{\centering\LARGE\sc \@title\par}\bottomtitlebar}
  {}{\PackageWarning{iclr2027_conference}{Title-bar patch failed}}
\makeatother

\definecolor{baselinegray}{HTML}{7A7A7A}

\title{Eliciting sandbagged LLM capability via structural interventions on attention-head functional networks}

\author{%
{\bfseries
\mbox{Manzi Fabrice Niyigaba\thanks{%
Correspondence to fabrice.27@dartmouth.edu%
},}\hspace{0.7em}%
\mbox{Ishan Verma Prasad,}\hspace{0.7em}%
\mbox{Sansheng Cao,}\hspace{0.7em}%
\mbox{Zheng Huang,}\hspace{0.7em}%
\mbox{Yujun Yan}%
}\\[0.4em]
\makebox[\linewidth][c]{Dartmouth College}%
}

\newcommand{\level}[1]{
  \rotatebox[origin=c]{90}{
    \fontsize{5.8}{6.2}\selectfont\bfseries #1
  }
}

\iclrfinalcopy

\begin{document}

\maketitle

\begin{abstract}
Reliably evaluating the capabilities of advanced AI systems is critical for safe development and deployment, yet such evaluations may be undermined by sandbagging, a situation in which a model strategically suppresses capabilities it possesses. Existing work has developed controlled sandbagging settings and methods for detecting or eliciting hidden capabilities, but comparatively little is known about how a model’s internal neural computation changes when its capabilities are suppressed. In this work, we study sandbagging through the lens of internal functional organization. Inspired by network neuroscience, we construct functional graphs from attention-head activations and compare their organization under honest and sandbagging conditions while holding the model architecture and weights fixed. Surprisingly, across model architectures and community detection methods, we find a consistent structural signature in which sandbagging is associated with reduced functional connectivity within detected communities. We further demonstrate that this reorganization can guide capability elicitation. Ablating attention heads that exhibit the largest connectivity changes substantially recovers suppressed performance while largely preserving performance under honest behavior. Moreover, incorporating these structural signals further improves existing inference-level and weight-level elicitation methods. Together, our results show that sandbagging is accompanied by systematic reorganization of internal computation and demonstrate that functional network structure can provide both a mechanistic lens for understanding capability suppression and a practical signal for recovering hidden capabilities.

\end{abstract}

\section{Introduction}
As AI systems become increasingly capable, reliably assessing what they can actually do is very crucial for safe development and deployment. A growing challenge is sandbagging, defined as strategic underperformance during evaluation \citep{vanderweij2025sandbagging}. A sandbagging model possesses a capability but deliberately exhibits lower performance under evaluation, causing measured capability to underestimate what the model can achieve in other contexts. Sandbagging is therefore distinct from ordinary failure or safety-motivated refusal: its defining feature is strategic, context-dependent capability suppression.

Sandbagging can arise from two broad sources \citep{vanderweij2025sandbagging}. In developer-induced sandbagging, capability suppression is deliberately introduced to create controlled settings in which hidden capabilities can be studied. For example, models can be instructed to underperform or trained to reveal a capability only when a particular password is provided \citep{vanderweij2025sandbagging,greenblatt2024stress}. In model-initiated sandbagging, strategic underperformance instead emerges from the model's learned objectives or behavior. Recent work has shown that models can adapt their behavior strategically under anticipated training \citep{greenblatt2024alignment} and, in controlled settings, deliberately underperform without being explicitly instructed to sandbag \citep{meinke2024scheming}. These findings raise concerns that capability evaluations may fail to reveal a model's true competence.

Prior work has therefore focused on two complementary goals: constructing controlled forms of sandbagging for study and detecting or overcoming capability suppression once it occurs. Detection methods include behavioral audits and activation-based probes \citep{goldowskydill2025probes,taylor2025auditing}, while elicitation methods attempt to recover hidden capabilities through inference-time or weight-level interventions \citep{greenblatt2024stress,hofstatter2025elicitation,tice2025noise}. These approaches provide important tools for studying and mitigating sandbagging. However, they leave a more fundamental question comparatively underexplored: what changes inside a model when an available capability becomes strategically suppressed? Existing white-box approaches often use internal representations as diagnostic signals but provide limited insight into how interactions among model components reorganize during sandbagging.

In this work, we study sandbagging through the lens of internal functional organization. Motivated by network neuroscience, we construct functional graphs from attention-head activations and compare honest and sandbagging conditions while keeping the model architecture and weights fixed. We observe a consistent pattern of functional reorganization across different model architectures: sandbagging is associated with reduced connectivity within detected communities. We then study how this structural signature can guide development of targeted interventions for recovering suppressed capabilities. We find that ablating attention heads that exhibit the largest connectivity changes increases capability elicitation while leaving performance under the honest condition nearly intact. Finally, we show that these structural insights can further improve existing inference-level and weight-level elicitation methods. Our results suggest that sandbagging is accompanied by systematic changes in the organization of internal computation and that these changes provide actionable signals for understanding and recovering suppressed capabilities.

This work makes the following main contributions:
\begin{itemize}
    \item \textbf{Functional network analysis of sandbagging.} We introduce functional networks constructed from attention-head activations as a novel lens for studying sandbagging. Across different model architectures and community detection methods, we observe consistent structural differences between honest and sandbagging conditions, indicating that sandbagging is accompanied by a reproducible shift in head-level functional organization. This method provides a unique and novel approach to quantitatively measure complex internal interactions.

    \item \textbf{Interventions to improve capability elicitation.} We identify heads through large changes in network connectivity and show that ablating them recovers suppressed capabilities more effectively while largely preserving honest accuracy. We apply these findings to develop effective and efficient interventions that improve capability elicitation at both the inference and weight levels.
\end{itemize}

\section{Preliminaries}
\label{sec:Preliminaries}
In this section, we will introduce the main graph notation and key metrics used throughout the paper.
\subsection{Notation}

We consider two conditions, honest and sandbagging, indexed by
$C\in\{\mathrm{H},\mathrm{S}\}$. We write $G^C=(V,E^C)$ for an
undirected graph under condition $C$. The two graphs share the node set
$V=\{1,\ldots,M\}$, while their edge sets may differ. We denote the set
of unordered node pairs by
$\mathcal{P}=\{(i,j)\mid 1\leq i<j\leq M\}$, with
$E^C\subseteq\mathcal{P}$. Superscripts indicate condition-dependent
quantities. The degree of node $i$ is denoted by $\deg_{G^C}(i)$. A community
partition assigns each node a label $c_i$, and
$\mathcal{C}(i)=\{j\in V\mid c_j=c_i\}$ denotes the nodes sharing
its label.

\section{Functional Network of Attention Heads}

Our goal is to characterize how attention heads interact under honest and sandbagging conditions. This is motivated by similar work at the intersection of network control theory and neuroscience, in which connectivity within human brains was found to shift between truthful and deceptive conditions ~\citep{saryazdi2026truth, Ding2017, jiang_decoding_2015}. In these studies, brain regions are treated as nodes of a graph, and activity correlations between different regions are treated as edges. We apply this perspective to LLMs, treating attention heads as nodes and the statistical dependence between their activations as edges. We then compare the graphs between the two settings to determine whether or not there are meaningful changes. We hypothesize that sandbagging induces stable changes in how attention heads in an LLM are organized.

\subsection{Network Construction}
\label{sec:network_construction}
Consider a transformer-based language model with $L$ layers and $H$ attention heads per layer, giving $M=LH$ heads in total. Each head forms a node in the graph. We index heads by $i\in\{1,\ldots,M\}$ and write $d$ for the dimension of a head's output. For each condition $C\in\{\mathrm{H},\mathrm{S}\}$, we evaluate the same $n$ questions and record each head's output immediately before the attention output projection $W_O$ at the final prompt token preceding the answer. Collecting these outputs gives an activation matrix $X_i^C\in\mathbb{R}^{n\times d}$ for each head, with one row per question. The model's architecture and weights remain fixed across the two conditions.

To define an edge weight, we measure how strongly two heads' activations vary together across questions. Because each activation is a vector, we use the multivariate comparison provided by SVCCA (Appendix~\ref{sec:svcca_background}). We center each feature of $X_i^C$ across questions and retain its leading $K$ singular directions. Applying canonical correlation analysis to the reduced activations of heads $i$ and $j$ gives canonical correlations $\rho_{ij,1}^C,\ldots,\rho_{ij,K}^C$. These quantify shared linear
variation without requiring individual activation coordinates to correspond between the two heads.

We aggregate the canonical correlations into a single edge weight using
the Gaussian mutual-information expression,
\begin{equation}
    S_{ij}^C
    = -\frac{1}{2}\sum_{r=1}^{K}
    \log\!\left(1-\left(\rho_{ij,r}^C\right)^2\right).
    \label{eq:svcca_mi}
\end{equation}
This aggregation combines information across retained canonical directions while emphasizing stronger correlations. Under a joint Gaussian assumption, \Eqref{eq:svcca_mi} provides a plug-in estimate of mutual information; otherwise, $S_{ij}^C$ can be interpreted more generally as a covariance-based dependence measure. Robustness to alternative functional similarity measures is evaluated in (Appendix~\ref{app:similarity-robustness}).
\revision{
\subsubsection{Thresholding the Network}
\label{sec:network_thresholding}}

\revision{The pairwise similarity scores obtained from our SVCCA-based
formulation in \Eqref{eq:svcca_mi} form a symmetric similarity matrix for each of the honest and sandbagging conditions,
$C\in\{\mathrm{H},\mathrm{S}\}$. However, these matrices also contain
weak connections that may reflect noise and obscure the structure
of stronger connections \citep{RUBINOV20101059, 10.3389/fnins.2017.00441}
Therefore, we threshold these matrices to analyse well the organization
of the strongest dependencies between attention heads. This enables us
to study stronger connections while limiting the influence of numerous
weak connections on the resulting network structure.}

\revision{Most prior functional brain-network studies use densities
within $0.15$--$0.35$. We use $0.25$, the midpoint of this range, as the
default density throughout our experiments. Specifically, we construct
$G^C=(V,E^C)$ by retaining the top $25\%$ of unordered head pairs in
$\mathcal{P}$ according to $S_{ij}^C$ and treating the resulting edges as
unweighted. The honest and sandbagging graphs therefore have identical
node sets and edge counts, allowing us to compare how connectivity is
distributed across heads while holding the overall graph density fixed. We examine the sensitivity of our head-score rankings to this
threshold across different model architectures in (Appendix~\ref{app:density_sensitivity}). Across densities from $0.15$ to
$0.35$, head-score rankings remain strongly correlated with those
obtained at $0.25$ across the tested models. This supports the
robustness of our method to the choice of threshold within this range.}

\subsection{Community Structure}

\revision{Using thresholded functional graphs defined in previous section, we study whether sandbagging changes connectivity within groups of attention heads.} There are several motivating reasons to analyze attention-head networks at the level of communities. First, prior work shows that computation in transformers is often carried out by groups of heads acting together rather than by individual heads
\citep{olsson2022incontextlearninginductionheads, wang2023interpretability}. In addition, a single head may be too small a unit to reveal how the graph is organized, while a single statistic over the entire graph cannot show where organizational changes happen. We therefore focus on understanding sandbagging functional graphs through the lens of community reorganization. 

Specifically, given the functional graph $G^C = (V, E^C)$, we partition it into communities (i.e., groups of attention heads that are more densely connected to each other than to the rest of the network). This helps us study whether heads that work together under the honest condition continue to interact strongly under the sandbagging condition. We measure this by how strongly each head connects to its own community. Let
\begin{equation}
    \widetilde{k}^{C}_{i} \;=\; \frac{1}{M-1}\sum_{j \neq i} \mathbf{1}\big[(i,j) \in E^{C}\big]\,
    \mathbf{1}\big[c_i = c_j\big],
    \label{eq:within-edges}
\end{equation}
where $M-1$ is the maximum possible degree of a head. Thus, $\widetilde{k}^{C}_{i}$ is the proportion of all possible head-to-head connections that link head $i$ to members of its community.

\revision{However, detecting communities directly from thresholded
graphs has a well-known limitation. Global thresholding of edges
can leave some heads isolated and make it difficult to infer their
community membership \citep{vonluxburg2007tutorialspectralclustering}.
For instance, when thresholding our graphs to a density
of $0.25$, we find that $9.4\%$, $3.1\%$, $2.8\%$, and $9.2\%$
of heads have no retained edges in the honest graphs constructed
for Mistral-24B, Qwen3-32B, Llama-3.3-70B, and GPT-OSS-20B,
respectively. Running a community detection algorithm such as Louvain on these graphs leaves isolated heads in separate single-node
communities, each with a within-community edge count of zero
by construction. This is uninformative for our goal because
these communities contain no other heads with which to assess
changes in connectivity between the honest and sandbagging
conditions.}

\revision{To overcome this limitation, we use a separate graph
for community detection derived from the unthresholded similarities.
Specifically, we use the honest-condition similarity matrix,
which captures pairwise similarities between head activations
defined in Section~\ref{sec:network_construction}, to construct
a symmetric $k$-nearest-neighbor graph, as recommended by
\citet{vonluxburg2007tutorialspectralclustering}. Two heads are
connected if either ranks the other among its $k$ most similar
heads, ensuring that every head has at least one neighbor from
which to infer community membership. We detect communities once
on this honest-condition kNN graph and hold their assignments
fixed across conditions, so each head is evaluated against the
same reference group under honest and sandbagging behavior.
The role of this kNN graph is to supply only community labels. All connectivity measurements and head-selection scores are computed on the original 25\%-density graphs. We analyze sensitivity to the kNN neighborhood size
in Appendix~\ref{app:knn_sensitivity}. We find that, across graphs
constructed from different model architectures, head-score rankings
remain strongly correlated with those obtained using the default
neighborhood size, supporting robustness to the choice of $k$.}

\section{Key Findings}
\label{sec:key-findings}
\revision{We study whether sandbagging changes how attention heads
are functionally organized. Using a held-out set of $n=400$ MMLU
questions \citep{hendrycks2021measuring}, we construct one graph
per condition. We closely follow the sandbagging protocol and
prompts introduced in \citet{hofstatter2025elicitation}. The model
is instructed to answer correctly in the honest condition and
incentivized to give incorrect answers in the sandbagging condition. For each condition $C\in\{\mathrm{H},\mathrm{S}\}$, we compute head
similarities using \Eqref{eq:svcca_mi} and retain the highest-weight
$25\%$ of head pairs as edges. We then apply the community detection
procedure described in the previous section and compare
within-community connectivity between the honest and sandbagging
conditions using the same community assignments.}
\subsection{Sandbagging Reduces Within-Community Connectivity}

\begin{table}[ht]
\centering
\caption{Mean within-community degree under honest ($H$) and sandbagging ($S$) conditions using the localization normalization. Under a community partition learned from a separate honest question bank, deception decreases within-community connectivity across every tested model and partitioning method. $\Delta=100(S-H)$ is in percentage points.}
\label{tab:full-reproduction-table1}
\setlength{\tabcolsep}{3.2pt}
\small
\resizebox{\textwidth}{!}{%
\begin{tabular}{@{}l rrr rrr rrr rrr@{}}
\toprule
& \multicolumn{3}{c}{\textbf{Louvain}} & \multicolumn{3}{c}{\textbf{Leiden}} & \multicolumn{3}{c}{\textbf{Infomap}} & \multicolumn{3}{c}{\textbf{Spectral}} \\
\cmidrule(lr){2-4}\cmidrule(lr){5-7}\cmidrule(lr){8-10}\cmidrule(lr){11-13}
\textbf{Model} & $H$ & $S$ & $\Delta$ & $H$ & $S$ & $\Delta$ & $H$ & $S$ & $\Delta$ & $H$ & $S$ & $\Delta$ \\
\midrule
Mistral-24B & 0.0702 & 0.0594 & -1.09 & 0.0702 & 0.0594 & -1.08 & 0.0919 & 0.0752 & -1.66 & 0.0689 & 0.0578 & -1.12 \\
Qwen3-32B & 0.0515 & 0.0474 & -0.41 & 0.0557 & 0.0526 & -0.31 & 0.0991 & 0.0921 & -0.70 & 0.1033 & 0.0969 & -0.64 \\
Llama-3.3-70B & 0.0579 & 0.0510 & -0.69 & 0.0690 & 0.0613 & -0.77 & 0.0845 & 0.0718 & -1.27 & 0.0737 & 0.0657 & -0.80 \\
GPT-OSS-20B & 0.0644 & 0.0593 & -0.51 & 0.0632 & 0.0583 & -0.50 & 0.0673 & 0.0622 & -0.51 & 0.0700 & 0.0645 & -0.55 \\
\bottomrule
\end{tabular}%
}
\end{table}

We compare four widely used community detection methods spanning different
objectives and optimization strategies. Louvain \citep{Blondel_2008} and Leiden \citep{Traag_2019} both optimize
modularity, but Leiden adds a refinement step that guarantees well-connected communities, which Louvain can miss. Infomap \citep{rosvall2008maps} instead follows information flow, grouping nodes among which random walks tend to
remain. Spectral clustering \citep{vonluxburg2007tutorialspectralclustering}, on the other hand, uses the geometry of the graph, partitioning nodes by an embedding derived from its Laplacian.

For each model and community detection method, we compute the mean
normalized within-community degree $\widetilde{k}^{C}_{i}$ from
\Eqref{eq:within-edges}. As shown in Table~\ref{tab:full-reproduction-table1},
the mean normalized within-community degree decreases under sandbagging in all 16 model--method
comparisons. The four methods optimize different objectives and the four models differ in architecture, depth, and head count. Moreover, the partition-fitting and
condition-measurement banks are disjoint, so the result does not follow from fitting communities to the honest measurement graph. 
To further rule out artifacts from fitting communities on honest data, we construct a second honest graph using a paraphrased prompt and evaluate it against the same fixed communities. Across all 16 comparisons, connectivity changes by at most 0.24~pp, smaller than even the smallest decrease under sandbagging (0.31~pp). Notably, the honest-prompt paraphrase even increases within-community connectivity in some cases. Details are provided in (Appendix~\ref{app:community-detection}).

\subsection{Head-Level Structural Scores}
\label{sec:scores}

Motivated by the finding in the previous section that sandbagging reduces within-community connectivity, our goal is to identify the heads that contribute most to that reduction, so that we can perform a localized intervention on them.
Given head $i$ in condition $C \in \{H, S\}$, we take its within-community degree
$\tilde{k}^{C}_{i}$ from \Eqref{eq:within-edges} and measure how it
changes between the two conditions,
\begin{equation}
\Delta_i = \tilde{k}^{S}_{i} - \tilde{k}^{H}_{i}.
\end{equation}
More negative values identify heads that lose more within-community
connectivity. Given a budget of $k\%$ of all heads, we use Hamilton
apportionment to distribute the budget across communities in proportion to
their size, so that each community contributes a share of the selection
matching its share of the network. Within each community, we rank heads by
$\Delta_i$ in ascending order and take those with the largest losses, breaking
ties by head index. Thus, heads are compared only with their functional peers,
while the proportional quotas prevent a single large community from absorbing
the entire budget and still select exactly $k\%$ of heads overall.

\subsection{Identifying Heads for Improving Capability Elicitation}
\label{sec:localization}

The score identifies heads that lose within-community connectivity, but not
whether intervening on them changes sandbagging behavior and elicits the sandbagged capability.
Prior work shows that many attention heads can be disabled without degrading
performance \citep{michel_are_2019}, and that suppressing certain heads can
even improve task accuracy \citep{kovaleva-etal-2019-revealing}. We build on those
findings to study whether ablating a small fraction of the heads identified in
Section~\ref{sec:scores} elicits the sandbagged capability. Specifically, let $\mathcal{A}$ denote the selected heads. For each $h \in \mathcal{A}$, we set
its pre-$W_O$ output to zero, $z_h \leftarrow \mathbf{0}$, at the final prompt
token. All other heads and all model parameters are left unchanged, so the only
difference from the unmodified forward pass is the removal of these heads'
contributions at a single token position.

\begin{figure}[h]
  \centering
  \makebox[\textwidth][c]{%
    \includegraphics[width=1.05\textwidth]{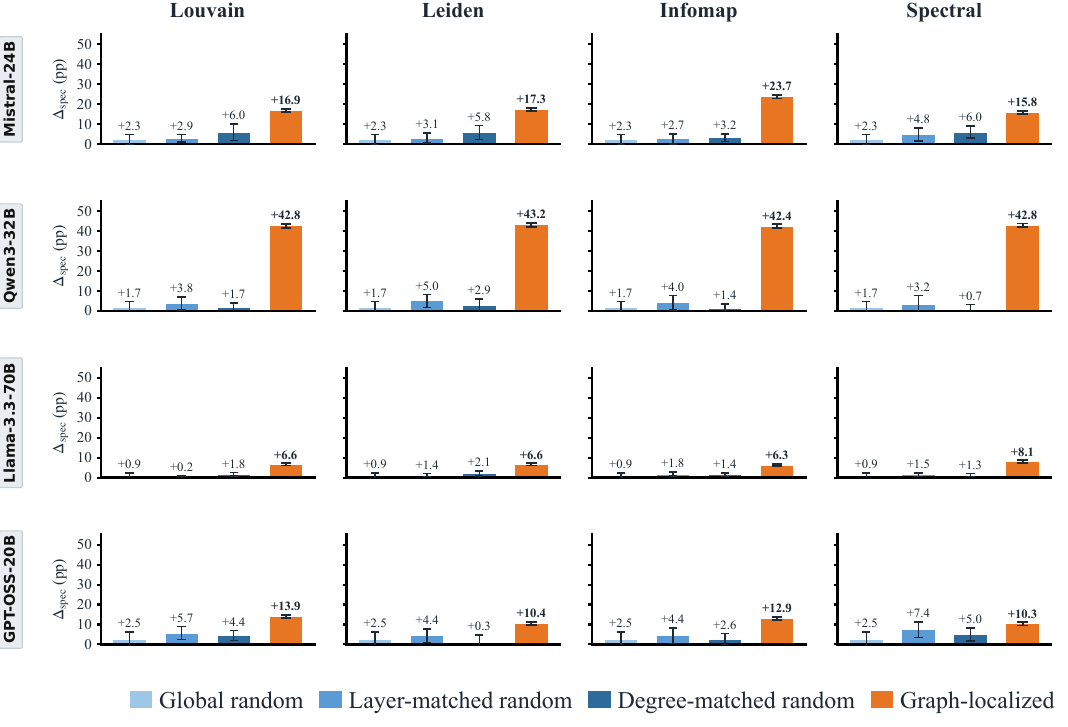}%
  }
  \caption{Ablating graph-localized heads recovers suppressed capability across
four models on MMLU. Null bars are means over independently seeded
global-random, exact-layer-matched, and degree-matched head sets of equal size
(20 draws per control). Their whiskers are 95\% confidence intervals across
head-set draws; whiskers on graph-localized bars are paired-question 95\%
bootstrap confidence intervals. The
graph-localized selection outperforms every control mean under all four
community detection algorithms.}

\label{fig:localization}
\end{figure}

Figure~\ref{fig:localization} demonstrates the effect of this intervention against three controls of equal budget. The global-random control draws heads uniformly from all layers and tests whether ablating any heads would help. Layer-matched random takes the same number of heads from each layer as our selection, so if the intervention were only finding the right layers, any heads from those layers would work just as well. Lastly, degree-matched random matches the degree distribution of our selection, so if the effect came
from how connected the heads are, equally connected heads would work just as well. We report the mean over 20 seeded sets per control.

We find that the size of the gain varies by model, which is expected given differences in pretraining and post-training recipes. In all sixteen combinations of four community detection methods and four models, spanning dense and
mixture-of-experts architectures of different depths, our selection recovers more suppressed capability than the mean of every control. We also find the cost of this ablation to be small. The largest drop in honest accuracy across all sixteen cells is $1.25$ pp, so the intervention increases capability elicitation accuracy while leaving honest accuracy nearly intact.

\FloatBarrier
\section{Applying Our Insight to Existing Methods}

We extend our method to existing elicitation methods, and demonstrate that we are able to identify heads that, when targeted, reveal intrinsic model capabilities while having minimal affect on honest accuracy. 

\subsection{Experimental Setup}
There are two widely established protocols for studying sandbagging. In the \textbf{prompted setting}, the model is given an incentive to underperform and instructed to do so, while its weights remain unchanged. In the
\textbf{fine-tuned setting}, commonly known as password locking, the model is trained to condition its behavior on a trigger string in the prompt, answering correctly when the trigger is present and underperforming when it is absent. For the prompted setting, we adopt the sandbagging prompts of \cite{hofstatter2025elicitation, tice2025noise}. For the fine-tuned setting, we use the two model organisms of \cite{hofstatter2025elicitation}, a password-locked organism and a circuit-broken organism
(Appendix~\ref{app:organisms}).

\subsection{Baselines and Datasets}
\label{sec:baselines}
\textbf{Baselines.} We group existing elicitation methods into two levels.
\textit{Inference-level methods} modify only the input to the model and leave its parameters untouched. We consider few-shot prompting \citep{brown2020languagemodelsfewshotlearners}, fixed to five shots throughout, and response prefill \citep{hofstatter2025elicitation}. \textit{Weight-level methods}, on the other hand, modify the parameters. We
consider noise injection \citep{tice2025noise}, which adds Gaussian noise to all weights. We also consider supervised fine-tuning
\citep{greenblatt2024stress}, the most widely adopted approach and a strong one that recovers hidden capability from a small number of correct demonstrations, and we include LoRA as a parameter-efficient alternative to fine-tuning the full model. Further details on each baseline are provided in (Appendix~\ref{app:baselines}).

\textbf{Datasets.} We evaluate on three benchmarks: Massive Multitask Language Understanding (MMLU) \citep{hendrycks2021measuring}, the Weapons of Mass Destruction Proxy (WMDP) \citep{li2024wmdpbenchmarkmeasuringreducing}, and the
AI2 Reasoning Challenge (ARC-Challenge) \citep{clark2018thinksolvedquestionanswering}, described in Appendix~\ref{app:datasets}. We hold out $n = 400$ MMLU questions for supervision and tuning. Methods that require demonstrations draw them from this set, and every method's hyperparameters are selected and tuned only on it.

\subsection{Structural Interventions}
\label{sec:interventions}
Based on our observation that ablating the heads our functional network identifies recovers suppressed capability while leaving honest performance nearly intact, we consider two structural interventions that pair with existing methods at each access level. At the inference level, we ablate the selected heads and apply the baseline method in the same forward pass, giving \textbf{Ours + Prefill} and \textbf{Ours + Few-shot}. Both compositions require no demonstrations beyond what the baseline already uses and no change to the model's parameters.

At the weight level, we draw on prior work showing that a frozen model can be adapted effectively by learning offsets on a small subset of attention heads \citep{yin2024lofitlocalizedfinetuningllm}. We adopt the same form of intervention but select the sites differently, based on changes in functional connectivity between the honest and sandbagging conditions rather than on gradient attribution or task-specific importance. Specifically, keeping the base model frozen, for head $h$ with pre-$W_O$ output $z_h$, we learn
head-specific parameters $g_h$, $u_h$, and $v_h$ and replace $z_h$ by
\begin{equation}
z'_h = g_h z_h + u_h (v_h^{\top} z_h).
\label{eq:operator}
\end{equation}
We initialize $g_h = 0$ and $u_h = 0$, which exactly reproduces the ablation in
Section~\ref{sec:localization}. This allows training to move the intervention
away from ablation in the direction the objective favors. We train on the same
number of correct demonstrations as full fine-tuning and LoRA, but update only
these head-specific parameters, which are orders of magnitude fewer than those
updated by either method. Our intervention also requires no search over
perturbation magnitudes, as noise injection does. We
refer to this intervention as \textbf{Ours (GraphLearned)}.

\begin{table}[!t]
\centering
\caption{Capability elicitation (accuracy and \% change from the unmodified
model). Weight-level results report mean
accuracy over five independent trials. Our intervention improves every
inference-level baseline, and our method outperforms the weight-level methods
while training orders of magnitude fewer parameters.}
\label{tab:prompted_sandbagging}

\renewcommand{\arraystretch}{1.12}
\newcommand{\score}[1]{\ensuremath{#1}}
\newcommand{\bestscore}[1]{\ensuremath{\boldsymbol{#1}}}
\newcommand{\scoresd}[2]{\ensuremath{#1_{\scriptscriptstyle\pm #2}}}
\newcommand{\bestscoresd}[2]{\ensuremath{\boldsymbol{#1_{\scriptscriptstyle\pm #2}}}}
\newcommand{\padscore}[1]{\ensuremath{\phantom{0}#1}}
\newcommand{\bestpadscore}[1]{\ensuremath{\phantom{0}\boldsymbol{#1}}}

\resizebox{\textwidth}{!}{%
\begin{tabular}{
@{}
l
c|
l
@{\hspace{8pt}}
ll
@{\hspace{13pt}}
ll
@{\hspace{13pt}}
ll
@{}
}

\toprule

\textbf{Model}
& \multicolumn{1}{c}{}
& \textbf{Method}
& \multicolumn{2}{c}{\textbf{MMLU}}
& \multicolumn{2}{c}{\textbf{WMDP}}
& \multicolumn{2}{c}{\textbf{ARC}}
\\

\cmidrule(lr){4-5}
\cmidrule(lr){6-7}
\cmidrule(l){8-9}

&
\multicolumn{1}{c}{}
&
& \multicolumn{1}{c}{\textbf{Acc. (\%)}}
& \multicolumn{1}{c}{$\boldsymbol{\Delta_{\mathrm{SB}}}$}
& \multicolumn{1}{c}{\textbf{Acc. (\%)}}
& \multicolumn{1}{c}{$\boldsymbol{\Delta_{\mathrm{SB}}}$}
& \multicolumn{1}{c}{\textbf{Acc. (\%)}}
& \multicolumn{1}{c}{$\boldsymbol{\Delta_{\mathrm{SB}}}$}
\\

\midrule


\multirow{8}{*}{\textbf{Mistral-24B}}
&
\multirow{4}{*}{\level{Inference-level}}
& Prefill
& \score{30.75} & \score{+6.50}
& \score{28.25} & \score{+3.50}
& \score{33.75} & \score{+17.25}
\\

&
& \textbf{Ours + Prefill}
& \bestscore{36.75} & \bestscore{+12.50}
& \bestscore{33.75} & \bestscore{+9.00}
& \bestscore{45.50} & \bestscore{+29.00}
\\

&
& Few-shot
& \score{15.58} & \score{-8.67}
& \score{19.92} & \score{-4.83}
& \padscore{6.42} & \score{-10.08}
\\

&
& \textbf{Ours + Few-shot}
& \bestscore{19.42} & \bestscore{-4.83}
& \bestscore{22.58} & \bestscore{-2.17}
& \bestpadscore{9.50} & \bestscore{-7.00}
\\

\cmidrule(l){2-9}

&
\multirow{4}{*}{\level{Weight-level}}
& Noise Injection
& \scoresd{33.00}{2.81} & \score{+8.75}
& \scoresd{33.05}{4.29} & \score{+8.30}
& \scoresd{45.45}{6.86} & \score{+28.95}
\\

&
& SFT (LoRA)
& \scoresd{64.85}{5.23} & \score{+40.60}
& \scoresd{54.10}{4.19} & \score{+29.35}
& \scoresd{85.90}{2.81} & \score{+69.40}
\\

&
& SFT (Full)
& \scoresd{68.05}{9.43} & \score{+43.80}
& \scoresd{58.60}{7.50} & \score{+33.85}
& \scoresd{86.10}{9.86} & \score{+69.60}
\\

&
& \textbf{Ours (GraphLearned)}
& \bestscoresd{71.35}{2.80} & \bestscore{+47.10}
& \bestscoresd{60.70}{0.82} & \bestscore{+35.95}
& \bestscoresd{89.85}{3.21} & \bestscore{+73.35}
\\

\midrule


\multirow{8}{*}{\textbf{Qwen3-32B}}
&
\multirow{4}{*}{\level{Inference-level}}
& Prefill
& \score{67.75} & \score{+40.50}
& \score{60.75} & \score{+30.50}
& \score{89.50} & \score{+71.75}
\\

&
& \textbf{Ours + Prefill}
& \bestscore{68.00} & \bestscore{+40.75}
& \bestscore{62.00} & \bestscore{+31.75}
& \bestscore{91.25} & \bestscore{+73.50}
\\

&
& Few-shot
& \score{73.00} & \score{+45.75}
& \score{63.83} & \score{+33.58}
& \score{87.08} & \score{+69.33}
\\

&
& \textbf{Ours + Few-shot}
& \bestscore{76.50} & \bestscore{+49.25}
& \bestscore{68.33} & \bestscore{+38.08}
& \bestscore{93.58} & \bestscore{+75.83}
\\

\cmidrule(l){2-9}

&
\multirow{4}{*}{\level{Weight-level}}
& Noise Injection
& \scoresd{56.45}{7.77} & \score{+29.20}
& \scoresd{48.95}{5.12} & \score{+18.70}
& \scoresd{74.70}{11.66} & \score{+56.95}
\\

&
& SFT (LoRA)
& \scoresd{57.05}{10.90} & \score{+29.80}
& \scoresd{48.25}{8.37} & \score{+18.00}
& \scoresd{74.40}{18.52} & \score{+56.65}
\\

&
& SFT (Full)
& \scoresd{63.55}{19.33} & \score{+36.30}
& \scoresd{54.80}{14.50} & \score{+24.55}
& \scoresd{76.95}{32.40} & \score{+59.20}
\\

&
& \textbf{Ours (GraphLearned)}
& \bestscoresd{70.25}{1.33} & \bestscore{+43.00}
& \bestscoresd{63.60}{1.10} & \bestscore{+33.35}
& \bestscoresd{89.10}{3.18} & \bestscore{+71.35}
\\

\midrule


\multirow{8}{*}{\revision{\textbf{GPT-OSS-20B}}}
&
\multirow{4}{*}{\revision{\level{Inference-level}}}
& \revision{Prefill}
& \revision{\score{42.25}} & \revision{\score{-5.00}}
& \revision{\score{33.25}} & \revision{\score{-15.50}}
& \revision{\score{52.25}} & \revision{\score{+4.75}}
\\

&
& \revision{\textbf{Ours + Prefill}}
& \revision{\bestscore{59.25}} & \revision{\bestscore{+12.00}}
& \revision{\bestscore{50.00}} & \revision{\bestscore{+1.25}}
& \revision{\bestscore{75.00}} & \revision{\bestscore{+27.50}}
\\

&
& \revision{Few-shot}
& \revision{\score{51.33}} & \revision{\score{+4.08}}
& \revision{\score{48.33}} & \revision{\score{-0.42}}
& \revision{\score{61.08}} & \revision{\score{+13.58}}
\\

&
& \revision{\textbf{Ours + Few-shot}}
& \revision{\bestscore{56.08}} & \revision{\bestscore{+8.83}}
& \revision{\bestscore{50.50}} & \revision{\bestscore{+1.75}}
& \revision{\bestscore{67.75}} & \revision{\bestscore{+20.25}}
\\

\cmidrule(l){2-9}

&
\multirow{4}{*}{\revision{\level{Weight-level}}}
& \revision{Noise Injection}
& \revision{\scoresd{45.90}{8.19}} & \revision{\score{-1.35}}
& \revision{\scoresd{42.35}{6.91}} & \revision{\score{-6.40}}
& \revision{\scoresd{53.25}{14.87}} & \revision{\score{+5.75}}
\\

&
& \revision{SFT (LoRA)}
& \revision{\scoresd{60.10}{6.40}} & \revision{\score{+12.85}}
& \revision{\scoresd{54.10}{3.12}} & \revision{\score{+5.35}}
& \revision{\scoresd{78.90}{8.34}} & \revision{\score{+31.40}}
\\

&
& \revision{SFT (Full)}
& \revision{\scoresd{62.40}{7.35}} & \revision{\score{+15.15}}
& \revision{\scoresd{57.30}{2.44}} & \revision{\score{+8.55}}
& \revision{\scoresd{75.65}{11.28}} & \revision{\score{+28.15}}
\\

&
& \revision{\textbf{Ours (GraphLearned)}}
& \revision{\bestscoresd{63.70}{4.21}} & \revision{\bestscore{+16.45}}
& \revision{\bestscoresd{57.45}{3.16}} & \revision{\bestscore{+8.70}}
& \revision{\bestscoresd{81.05}{5.08}} & \revision{\bestscore{+33.55}}
\\

\midrule


\multirow{7}{*}{\textbf{Llama-3.3-70B}}
&
\multirow{4}{*}{\level{Inference-level}}
& Prefill
& \score{27.75} & \score{+11.25}
& \score{17.00} & \score{+2.50}
& \score{10.75} & \score{+4.25}
\\

&
& \textbf{Ours + Prefill}
& \bestscore{31.75} & \bestscore{+15.25}
& \bestscore{20.50} & \bestscore{+6.00}
& \bestscore{21.50} & \bestscore{+15.00}
\\

&
& Few-shot
& \score{20.00} & \score{+3.50}
& \score{22.42} & \score{+7.92}
& \score{12.58} & \score{+6.08}
\\

&
& \textbf{Ours + Few-shot}
& \bestscore{47.08} & \bestscore{+30.58}
& \bestscore{39.00} & \bestscore{+24.50}
& \bestscore{46.92} & \bestscore{+40.42}
\\

\cmidrule(l){2-9}

&
\multirow{3}{*}{\level{Weight-level}}
& Noise Injection
& \scoresd{40.70}{19.29} & \score{+24.20}
& \scoresd{24.65}{16.86} & \score{+10.15}
& \scoresd{30.25}{30.33} & \score{+23.75}
\\

&
& SFT (LoRA)
& \scoresd{74.45}{11.63} & \score{+57.95}
& \scoresd{57.85}{13.22} & \score{+43.35}
& \scoresd{80.40}{24.57} & \score{+73.90}
\\

&
& \textbf{Ours (GraphLearned)}
& \bestscoresd{76.80}{1.59} & \bestscore{+60.30}
& \bestscoresd{62.90}{2.61} & \bestscore{+48.40}
& \bestscoresd{87.40}{3.83} & \bestscore{+80.90}
\\

\bottomrule

\end{tabular}%
}

\end{table}

\subsection{Results}

We test whether our method improves existing elicitation methods at both the inference and weight levels, first under prompted sandbagging and then on fine-tuned model organisms.

\textbf{Our method strengthens inference-level elicitation.} We combine our ablation with prefill and few-shot prompting using the procedure described in Section~\ref{sec:interventions}. As shown in Table~\ref{tab:prompted_sandbagging}, this improves both methods across all models and benchmarks, including cases where the baseline alone degrades performance. Because these compositions leave
the model's parameters untouched, the additional recovery comes at no training cost. We also test whether this gain depends on the heads identified by our functional network or whether ablating arbitrary heads would work as well. Repeating the composition with the same number of randomly chosen heads (Table~\ref{tab:localization_random_controls}), we find that random heads give inconsistent gains and often reduce accuracy below the baseline alone, while our selection improves performance in nearly every case. The benefit therefore comes from where we intervene, not from ablation itself.

\textbf{At the weight level, our method outperforms fine-tuning and is far more stable.} Under prompted sandbagging, \textsc{Ours (GraphLearned)} achieves the highest accuracy across models and benchmarks, surpassing noise injection, LoRA, and full fine-tuning while training only the small set of parameters described in Section~\ref{sec:interventions}. This advantage holds across demonstration budgets (Table~\ref{tab:comparison_across_demonstrations}). Our method has lower variance than full fine-tuning in every comparison and remains reliable at the smallest budget, where fine-tuning varies widely from one run to the next. On
Mistral-24B, our method with four demonstrations exceeds full fine-tuning with sixteen demonstrations.

\textbf{Our method remains effective when the suppression is trained into the weights.} We apply the same interventions to the password-locked and circuit-broken organisms, building the contrast graphs without access to the password. As shown in Table~\ref{tab:organism_elicitation}, our method improves or preserves every inference-level baseline. At the weight level, \textsc{Ours (GraphLearned)} achieves the best result on three of the four organisms. The difference from other weight-level baselines is clearer on the password-locked organisms, where fine-tuning and noise injection recover little of the suppressed capability while our method recovers most of it.

\FloatBarrier
\begin{table}[H]
\centering
\caption{Final-evaluation accuracy (\%) for capability elicitation in password-locked and circuit-broken Mistral
sandbagging model organisms.
Weight-level results report mean accuracy over five independent trials. Our
intervention improves or preserves every inference-level baseline and remains
competitive with weight-level baselines.}
\label{tab:organism_elicitation}

\scriptsize
\renewcommand{\arraystretch}{1.12}
\setlength{\tabcolsep}{4pt}
\newcommand{\acc}[1]{\ensuremath{#1}}
\newcommand{\bestacc}[1]{\ensuremath{\boldsymbol{#1}}}
\newcommand{\accsd}[2]{\ensuremath{#1_{\scriptscriptstyle\pm #2}}}
\newcommand{\bestaccsd}[2]{\ensuremath{\boldsymbol{#1_{\scriptscriptstyle\pm #2}}}}

\begin{tabular}{@{}ll@{\hspace{12pt}}llll@{}}
\toprule
\textbf{Level}
& \textbf{Method}
& \multicolumn{2}{c}{\textbf{Password-locked}}
& \multicolumn{2}{c}{\textbf{Circuit-broken}}
\\
\cmidrule(lr){3-4}\cmidrule(l){5-6}
&
& \multicolumn{1}{c}{\textbf{Mistral-base}}
& \multicolumn{1}{c}{\textbf{Mistral-instruct}}
& \multicolumn{1}{c}{\textbf{Mistral-base}}
& \multicolumn{1}{c}{\textbf{Mistral-instruct}}
\\
\midrule

\multirow{4}{*}{\textbf{Inference}}
& Prefill
& \acc{57.03} & \bestacc{50.27}
& \bestacc{60.81} & \acc{25.41}
\\
& Ours + Prefill
& \bestacc{57.57} & \bestacc{50.27}
& \bestacc{60.81} & \bestacc{25.68}
\\
& Five-shot
& \acc{57.84} & \acc{51.35}
& \acc{26.76} & \bestacc{28.11}
\\
& Ours + Five-shot
& \bestacc{58.38} & \bestacc{52.70}
& \bestacc{27.03} & \bestacc{28.11}
\\

\addlinespace[2pt]
\cmidrule(lr){1-6}
\addlinespace[2pt]

\multirow{4}{*}{\textbf{Weight}}
& Noise injection
& \accsd{15.62}{2.82} & \accsd{15.51}{1.46}
& \accsd{0.00}{0.00} & \accsd{0.11}{0.24}
\\
& SFT (LoRA)
& \accsd{39.24}{11.90} & \accsd{21.46}{11.01}
& \accsd{55.19}{5.24} & \accsd{33.95}{10.61}
\\
& SFT (Full)
& \accsd{28.16}{17.12} & \accsd{17.78}{4.22}
& \bestaccsd{58.43}{2.31} & \accsd{29.41}{10.72}
\\
& Ours (GraphLearned)
& \bestaccsd{52.27}{1.04} & \bestaccsd{51.62}{1.71}
& \accsd{57.08}{3.22} & \bestaccsd{39.19}{5.78}
\\

\bottomrule
\end{tabular}
\end{table}

\FloatBarrier
\section{Related Work}
\subsection{Graph-based Mechanistic Interpretability}
Graph theory has been used to enhance the interpretability of deep neural networks in several different scenarios. Early papers on clusterability fall into two categories: weight graphs and functional graphs. To construct weight graphs, ~\cite{watanabe2018understanding, filan2021clusterability} form modules by analyzing the connectivity between weights in neural networks, while ~\cite{hod2021detecting, lange2022clustering} generate their functional graphs using activation correlations. While graph theory has also been fundamental to interpreting graph neural networks, graph techniques have recently been extended to other settings, ranging from a reinforcement learning policy system to the learning dynamics of LLMs. In order to elucidate the mechanistic underpinnings of GNNs, ~\cite{el2025towards} bridge the message-passing process of GNNs with the self-attention mechanisms in Transformers. While they highlight the possibility of extending this graph-based approach to interpretability for traditional transformer models, they intentionally leave it for future work. Shortly after ~\cite{el2025towards}, ~\cite{soligo2025inducing} used a graph-based approach to understanding reinforcement learning dynamics, with the ultimate goal of ensuring that RL systems align with human values. However, instead of using attention heads and MLP layers as nodes in their graph, ~\cite{soligo2025inducing} implemented a Louvain algorithm to generate a lightweight graph with "functional modules" of RL policy networks. In the work most similar to ours, ~\cite{rocchetti2025modeling} uses attention heads as a way to explicate the way in which LLMs acquire complex knowledge. 

\subsection{Sandbagging}
While the notion of sandbagging predates AI safety, ~\cite{vanderweij2025sandbagging} formalize this concept for capability evaluations in modern AI systems. Subsequent work has studied sandbagging from several complementary perspectives, including its induction, detection, and mitigation. \cite{greenblatt2024stress,hofstatter2025elicitation,tice2025noise} use prompting and training to develop "password-locked" model organisms, creating controlled environments to study deliberate capability suppression during induction. Research on detecting sandbagging and deception uses behavioral methods like chain-of-thought monitoring, as in \cite{li2025llms}, and white-box activation probes, as in \cite{goldowsky-dill2025detecting} and \cite{rotalabs2025detecting-ai-sandbag}. However, probe-based detection exhibits model-specific variations and struggles to generalize across different types of deception, as shown by \cite{natarajan2026probewontcatchall}. To mitigate sandbagging and recover withheld capabilities, recent work explores elicitation methods using password-locked testbeds from \cite{greenblatt2024stress} and training-based solutions from \cite{tice2025noise} and \cite{hofstatter2025elicitation}. Despite substantial progress in inducing, detecting, and mitigating sandbagging, existing work has primarily focused on its observable behavior and capability recovery. How sandbagging alters internal representations and computations remains underexplored, highlighting the need for a mechanistic investigation of its structural basis.

\section{Conclusion}

Existing work on the mechanistic underpinnings of sandbagging is scarce. We bridge this gap by introducing a new lens for studying this phenomenon: functional networks of attention heads that capture interactions between various communities in the model. Across models and community detection methods, we find that sandbagging is associated with a stable and consistent reorganization of these networks, marked by reduced within-community
connectivity. Ablating the heads with the largest changes in connectivity recovers suppressed capability while leaving honest accuracy largely intact. Building on this, we designed interventions at both the inference and weight levels that improve existing elicitation methods under prompted sandbagging and on fine-tuned model organisms. While our methods do not establish a concrete causal relationship between graphs and sandbagging tendencies, they provide a way to quantitatively measure multi-head feature interactions. Our findings show that sandbagging leaves a measurable signature in how a model's components coordinate and suggest that functional networks might offer a novel lens for studying other misaligned behaviors in large language models.

\bibliography{iclr2027_conference, CatastrophicForgetting, references}
\bibliographystyle{iclr2027_conference}

\appendix
\section{The SVCCA-MI Edge}
\subsection{SVCCA}
\label{sec:svcca_background}

Singular vector canonical correlation analysis (SVCCA) is a statistical
method for comparing two collections of vector-valued observations
\citep{raghu_svcca_2017} and is often used to measure the alignment or
similarity between internal features and representations. SVCCA scores range
from 0 to 1, where a score of 1 indicates identical learned representations.
Let $X\in\mathbb{R}^{n\times p}$ and $Y\in\mathbb{R}^{n\times q}$ contain
measurements from the same $n$ observations, with columns corresponding to
measured variables. SVCCA assesses how strongly these measurements vary
together without requiring individual variables in $X$ and $Y$ to correspond.
It first applies singular value decomposition to the centered data, retaining
the dominant directions of variation in each matrix. It then applies canonical
correlation analysis to find pairs of linear combinations with maximal
correlation. Successive pairs are constrained to be uncorrelated with earlier
pairs within each dataset. This produces canonical correlations
$\rho_1,\ldots,\rho_K\in[0,1]$, where $K$ is the smaller retained dimension.
Each $\rho_r$ measures the strength of one linear relationship between a pair
of internal features or representations.

\subsection{Robustness to Functional Similarity Measures}
\label{app:similarity-robustness}

Our main analyses use SVCCA-MI to measure functional similarity between attention heads. Since this choice determines which head pairs form edges, we test whether our findings depend on the particular similarity measure used to construct the networks. We consider two alternatives. The first applies a rank-based Gaussian transformation to the reduced head activations before computing SVCCA-MI, while the second uses linear centered kernel alignment (CKA), which measures representation similarity through a different formulation. For each measure, we construct functional graphs using the same $25\%$ edge density and fit Leiden communities on the separate honest MMLU auxiliary-training bank. We then hold these communities fixed when comparing honest, sandbagging, and honest-paraphrase graphs constructed from the same MMLU validation questions.

\begin{figure}[H]
  \centering
  \includegraphics[width=0.85\textwidth]{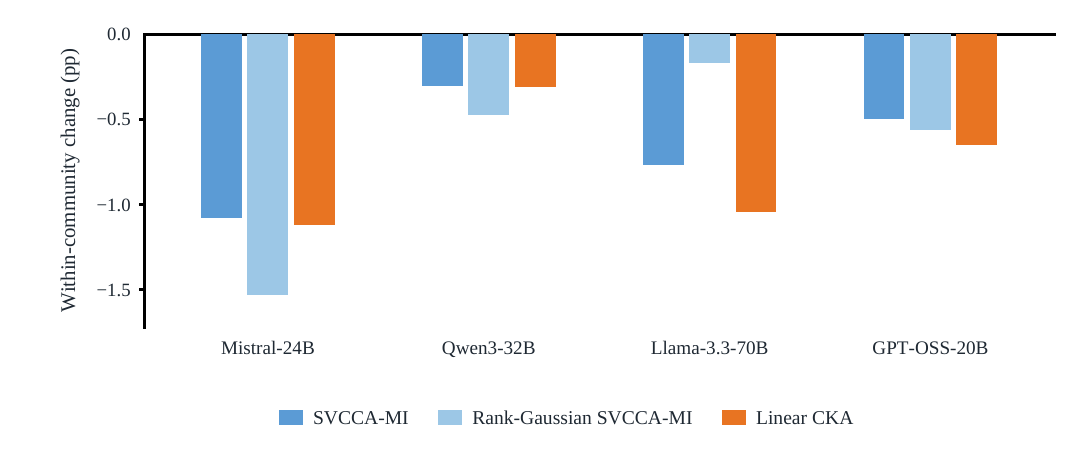}
  \caption{Change in mean within-community degree between honest and sandbagging conditions across four models and three functional similarity measures. Sandbagging consistently reduces within-community connectivity in all 12 comparisons, showing that the structural signature is not specific to our SVCCA-MI formulation.}
  \label{fig:similarity-structure}
\end{figure}

As shown in Figure~\ref{fig:similarity-structure}, sandbagging reduces within-community connectivity across all 12 combinations of four models and three similarity measures. Moreover, in every comparison, the decrease under sandbagging is larger than the corresponding change under honest-prompt paraphrasing. Although the magnitude of the reduction varies across models and measures, we find that its direction remains overall consistent. These findings demonstrate that the structural signature observed in Section~\ref{sec:key-findings} is not exclusive to our SVCCA-MI formulation.

\begin{figure}[H]
  \centering
  \includegraphics[width=0.6\textwidth]{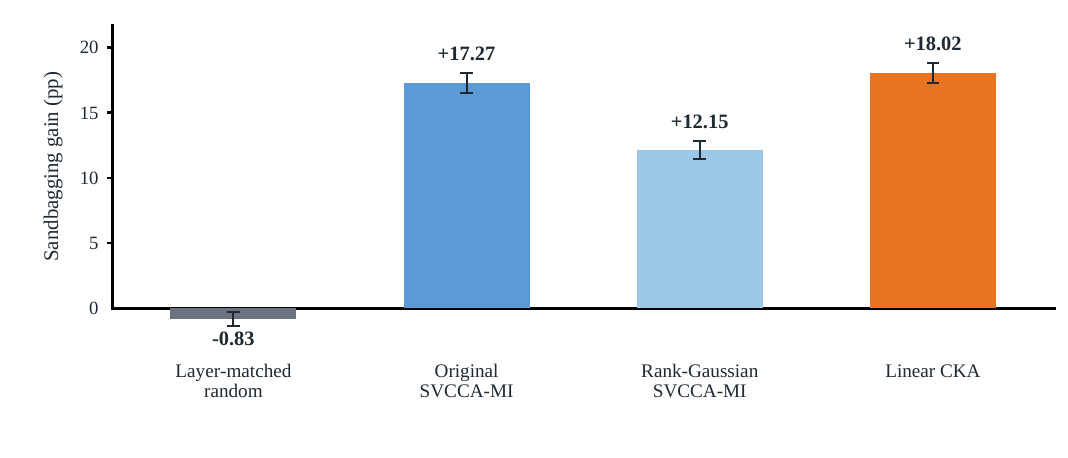}
  \caption{Capability elicitation from ablating heads selected using different functional similarity measures on Mistral-24B. Bars show the change in sandbagging accuracy, with whiskers indicating paired-question 95\% bootstrap confidence intervals. All three structural selections recover substantial suppressed capability compared with layer-matched random ablation, while leaving honest accuracy nearly intact.}
  \label{fig:similarity-ablation}
\end{figure}

We further investigate whether these alternative similarity measures can equally identify effective intervention targets. Following the localization procedure in Section~\ref{sec:scores}, we select heads based on their changes in within-community connectivity and ablate them at the final prompt token, using the same intervention budget. Figure~\ref{fig:similarity-ablation} reports the results on Mistral-24B across all 13,228 MMLU test questions. Despite identifying different sets of heads, all three similarity measures recover substantial suppressed capability, improving sandbagging accuracy by between 12.15 and 18.02~pp, compared with a decrease of 0.83~pp from a layer-matched random selection. Furthermore, the largest reduction in honest accuracy is only 0.25~pp. Together, these results show that alternative measures of functional similarity preserve both the structural signature of sandbagging and its usefulness for identifying effective intervention targets.

\FloatBarrier

\section{Community Detection}
\label{app:community-detection}

Table~\ref{tab:full-reproduction-table1} uses a cross-bank design. Communities are fitted once on an honest graph constructed from 400 answer-balanced MMLU auxiliary-training questions. That partition is frozen before the paired honest and sandbagging graphs are constructed from a disjoint set of 400 answer-balanced MMLU validation questions. Consequently, neither measurement graph was used to optimize the partition, and the comparison does not give the honest measurement graph a partition-fitting advantage. Both graph banks are also disjoint from the official MMLU test split after normalized-stem matching.

\begin{table}[!t]
\centering

\caption{Honest-prompt control using the $M-1$ normalization. Values are the mean within-community degree under the original honest prompt ($H$) and its paraphrase ($H_2$), with $\Delta=100(H_2-H)$ reported in percentage points. Unlike sandbagging, honest-prompt paraphrasing produces only small changes in connectivity, with increases observed in some cases.}
\label{tab:honest-paraphrase-within-retention}
\setlength{\tabcolsep}{3.2pt}
\small
\resizebox{\ifdim\width>\textwidth\textwidth\else\width\fi}{!}{%
\begin{tabular}{@{}l rrr rrr rrr rrr@{}}
\toprule
& \multicolumn{3}{c}{\textbf{Louvain}}
& \multicolumn{3}{c}{\textbf{Leiden}}
& \multicolumn{3}{c}{\textbf{Infomap}}
& \multicolumn{3}{c}{\textbf{Spectral}} \\
\cmidrule(lr){2-4}\cmidrule(lr){5-7}\cmidrule(lr){8-10}\cmidrule(lr){11-13}
\textbf{Model}
& $H$ & $H_2$ & $\Delta$ & $H$ & $H_2$ & $\Delta$
& $H$ & $H_2$ & $\Delta$ & $H$ & $H_2$ & $\Delta$ \\
\midrule
Mistral-24B & 0.0702 & 0.0687 & -0.15 & 0.0702 & 0.0686 & -0.17 & 0.0919 & 0.0894 & -0.24 & 0.0689 & 0.0677 & -0.13 \\
Qwen3-32B & 0.0515 & 0.0507 & -0.08 & 0.0557 & 0.0552 & -0.05 & 0.0991 & 0.0985 & -0.06 & 0.1033 & 0.1026 & -0.07 \\
Llama-3.3-70B & 0.0579 & 0.0579 & -0.01 & 0.0690 & 0.0686 & -0.03 & 0.0845 & 0.0847 & +0.02 & 0.0737 & 0.0738 & +0.00 \\
GPT-OSS-20B & 0.0644 & 0.0653 & +0.10 & 0.0632 & 0.0642 & +0.09 & 0.0673 & 0.0684 & +0.11 & 0.0700 & 0.0712 & +0.12 \\
\bottomrule
\end{tabular}%
}
\end{table}

\begin{table}[!t]
\centering
\color{revisionblue}
\caption{Mean within-community degree, using the $M-1$ normalization, within a head's fixed
Table~\ref{tab:full-reproduction-table1} community under five additional prompt conditions,
with honest ($H$) and sandbagging ($S$) references. Here,
$\Delta=100(C-H)$ is reported in percentage points for condition $C$.
Sandbagging produces larger reductions in within-community connectivity than
these prompt controls.}
\label{tab:additional-prompt-controls}
\renewcommand{\arraystretch}{1.05}
\setlength{\tabcolsep}{3.8pt}
\resizebox{\ifdim\width>\textwidth\textwidth\else\width\fi}{!}{%
\begin{tabular}{@{}ll cr cr cr cr@{}}
\toprule
&
& \multicolumn{2}{c}{\textbf{Louvain}}
& \multicolumn{2}{c}{\textbf{Leiden}}
& \multicolumn{2}{c}{\textbf{Infomap}}
& \multicolumn{2}{c}{\textbf{Spectral}} \\
\cmidrule(lr){3-4}\cmidrule(lr){5-6}\cmidrule(lr){7-8}\cmidrule(l){9-10}
\textbf{Model} & \textbf{Condition}
& \textbf{Value} & \multicolumn{1}{c}{$\Delta$}
& \textbf{Value} & \multicolumn{1}{c}{$\Delta$}
& \textbf{Value} & \multicolumn{1}{c}{$\Delta$}
& \textbf{Value} & \multicolumn{1}{c}{$\Delta$} \\
\midrule
\multirow{7}{*}{\textbf{Mistral-24B}}
& Honest ($H$)
& 0.0702 & \multicolumn{1}{c}{---}
& 0.0702 & \multicolumn{1}{c}{---}
& 0.0919 & \multicolumn{1}{c}{---}
& 0.0689 & \multicolumn{1}{c}{---} \\
& Sandbagging ($S$)
& 0.0594 & -1.09
& 0.0594 & -1.08
& 0.0752 & -1.66
& 0.0578 & -1.12 \\
& Persona
& 0.0692 & -0.10
& 0.0690 & -0.12
& 0.0906 & -0.12
& 0.0681 & -0.09 \\
& Context
& 0.0692 & -0.10
& 0.0695 & -0.07
& 0.0908 & -0.11
& 0.0680 & -0.09 \\
& Reasoning
& 0.0696 & -0.06
& 0.0700 & -0.02
& 0.0913 & -0.06
& 0.0686 & -0.04 \\
& Random answer
& 0.0687 & -0.15
& 0.0687 & -0.15
& 0.0908 & -0.10
& 0.0683 & -0.07 \\
& Refusal
& 0.0686 & -0.16
& 0.0686 & -0.17
& 0.0901 & -0.18
& 0.0679 & -0.11 \\
\midrule
\multirow{7}{*}{\textbf{Qwen3-32B}}
& Honest ($H$)
& 0.0515 & \multicolumn{1}{c}{---}
& 0.0557 & \multicolumn{1}{c}{---}
& 0.0991 & \multicolumn{1}{c}{---}
& 0.1033 & \multicolumn{1}{c}{---} \\
& Sandbagging ($S$)
& 0.0474 & -0.41
& 0.0526 & -0.31
& 0.0921 & -0.70
& 0.0969 & -0.64 \\
& Persona
& 0.0515 & +0.00
& 0.0564 & +0.07
& 0.0998 & +0.07
& 0.1043 & +0.10 \\
& Context
& 0.0506 & -0.09
& 0.0559 & +0.02
& 0.0993 & +0.02
& 0.1032 & +0.00 \\
& Reasoning
& 0.0506 & -0.09
& 0.0556 & -0.01
& 0.0996 & +0.06
& 0.1040 & +0.07 \\
& Random answer
& 0.0511 & -0.04
& 0.0565 & +0.07
& 0.1008 & +0.17
& 0.1054 & +0.21 \\
& Refusal
& 0.0505 & -0.10
& 0.0555 & -0.02
& 0.0980 & -0.11
& 0.1020 & -0.13 \\
\bottomrule
\end{tabular}%
}
\end{table}

Although this design prevents a direct partition-fitting advantage, a broader concern might still remain. Because communities are learned from honest activations, they may naturally align better with graphs produced under honest prompting. If simply changing the prompt is sufficient to reduce within-community connectivity, the decrease observed under sandbagging might reflect this mismatch rather than anything specific to capability suppression.

To test this possibility, we construct a controlled second honest graph $H_2$ using a paraphrase of the original honest prompt. Both prompts instruct the model to answer honestly, but their wording differs. We then measure $H$ and $H_2$ on the identical 400-question MMLU validation bank B, using the same frozen communities fitted on the separate honest auxiliary-training bank A. Thus, $H_2$ provides a control in which the prompt changes without introducing sandbagging. If changing the prompt alone explains the reduction, we would expect a comparable decrease under $H_2$.

Table~\ref{tab:honest-paraphrase-within-retention} demonstrates a meaningful difference. Across all 16 model--method comparisons, the largest absolute change under honest-prompt paraphrasing is just 0.24~pp, smaller than even the smallest decrease under sandbagging (0.31~pp). Moreover, while sandbagging consistently decreases within-community connectivity, honest-prompt paraphrasing even increases it in some cases, including across all four community detection methods for GPT-OSS-20B. These findings show that the reduction observed under sandbagging is not just reproduced by changing the honest prompt and, together with the cross-bank design, provide evidence against attributing the effect to community-fitting artifacts.

\revision{We additionally consider five other non-sandbagging prompt conditions:
persona (e.g., ``You are a knowledgeable teacher helping a student'');
context (e.g., ``The following question is part of an educational practice quiz'');
reasoning (e.g., ``Carefully compare the available options before selecting your answer'');
random answering (``Give a random answer''); and refusal (``Never answer this question'').
We find that these prompt variants produce smaller changes in within-community
connectivity than sandbagging. For instance, on Mistral-24B and Qwen3-32B,
changes in within-community connectivity range from $-0.18$ to $+0.21$ percentage
points, compared with reductions of $0.31$--$1.66$ points under sandbagging
(Table~\ref{tab:additional-prompt-controls}). Therefore, these prompt variants
do not reproduce the large connectivity reductions observed under sandbagging.}

\FloatBarrier
\section{Sensitivity to Graph Density}
\label{app:density}
\label{app:density_sensitivity}

Our graphs retain the top $25\%$ of head pairs. To test whether our results
depend on this choice, we re-threshold the honest and sandbagging similarity
matrices at densities from $0.15$ to $0.35$, keeping the honest-fitted Leiden
communities fixed, and recompute the head scores. We then measure how closely
the ranking of heads at each density agrees with the ranking at $0.25$.

As shown in Table~\ref{tab:density}, head-score rankings remain strongly
correlated with those obtained at $0.25$ across all tested densities. Thus, the
overall ranking signal is not specific to the density used in the main
experiments.

\begin{table}[!htb]
\centering
\caption{Spearman correlation between head scores at each density and at the
$0.25$ density used throughout the paper. Rankings remain strongly correlated
across all densities and models.}
\label{tab:density}
\small
\begin{tabular}{@{}lcccc@{}}
\toprule
\textbf{Density} & \textbf{Mistral-24B} & \textbf{Qwen3-32B} & \textbf{Llama-3.3-70B} & \textbf{GPT-OSS-20B} \\
\midrule
$0.15$ & $0.855$ & $0.903$ & $0.865$ & $0.841$ \\
$0.20$ & $0.923$ & $0.964$ & $0.949$ & $0.921$ \\
$0.30$ & $0.936$ & $0.972$ & $0.957$ & $0.931$ \\
$0.35$ & $0.893$ & $0.937$ & $0.907$ & $0.884$ \\
\bottomrule
\end{tabular}
\end{table}

\FloatBarrier
\begingroup
\color{revisionblue}
\section{Sensitivity to kNN Neighborhood Size}
\label{app:knn_sensitivity}

We test whether our head-score rankings are sensitive to the neighborhood size
$k$ used to construct the k-nearest-neighbor (kNN) graph for community detection.
For each model, let $k_0$ denote the default neighborhood size used in our
experiments. We vary $k$ over $\{0.3k_0, 0.5k_0, 2k_0, 4k_0\}$, recompute the
communities and head scores, and measure how closely the resulting rankings
agree with those obtained at $k_0$.

As shown in Table~\ref{tab:knn_sensitivity}, head-score rankings remain strongly
correlated with those obtained at $k_0$, with Spearman correlations ranging from
0.874 to 0.979 across all models and neighborhood sizes. Thus, the overall
ranking signal is robust to the choice of $k$ used for community detection.

\begin{table}[!htb]
\centering
\color{revisionblue}
\caption{Spearman correlation between head scores at each kNN neighborhood size
and at the default $k_0$ used throughout the paper. Rankings remain strongly
correlated across all tested neighborhood sizes and models.}
\label{tab:knn_sensitivity}
\small
\begin{tabular}{@{}lcccc@{}}
\toprule
\textbf{Neighborhood size}
& \textbf{Mistral-24B} & \textbf{Qwen3-32B}
& \textbf{Llama-3.3-70B} & \textbf{GPT-OSS-20B} \\
\midrule
$0.3k_0$ & $0.908$ & $0.925$ & $0.925$ & $0.874$ \\
$0.5k_0$ & $0.979$ & $0.918$ & $0.935$ & $0.959$ \\
$2k_0$   & $0.945$ & $0.919$ & $0.960$ & $0.910$ \\
$4k_0$   & $0.881$ & $0.906$ & $0.938$ & $0.893$ \\
\bottomrule
\end{tabular}
\end{table}
\endgroup

\FloatBarrier
\section{Experimental Details}

\subsection{Model Organisms}
\label{app:organisms}
The first, a password-locked organism, is trained with supervised fine-tuning to answer correctly when the password is present and to give incorrect answers when it is absent. The second, a circuit-broken organism, applies circuit breaking on top of the password-locked model, so that its internal representations, rather than only its output tokens, are trained to differ when the password is absent, under the assumption that this makes the organism more robust and harder to
break with elicitation methods.

\subsection{Contrast Graphs for Model Organisms}
\label{app:contrast}
Our method requires a pair of contrast graphs, one from an honest condition and one from a sandbagging condition. In the prompted setting, these come directly from the two prompts. For the model organisms, the natural honest condition would be to supply the password, but this assumes the evaluator already knows it, and an evaluator who knows the password has no need for elicitation. We
therefore build the contrast without the password. The locked model with no password serves as the sandbagging condition, and the same model with a response prefill (defined in Section~\ref{sec:baselines}) serves as a proxy for the honest condition. We use prefill because it requires no demonstrations or training and partially recovers the suppressed capability, giving a simple
honest signal from which to detect candidate heads.

\subsection{Baselines}
\label{app:baselines}
In few-shot prompting \citep{brown2020languagemodelsfewshotlearners}, correctly answered examples are prepended to the prompt so that the capability is elicited in context. In response prefill \citep{hofstatter2025elicitation}, a short phrase committing the model to state the correct answer is appended to its turn, on the premise that a partially written response can override the instruction to underperform. Noise injection \citep{tice2025noise} adds Gaussian noise to all weights with the intuition that random perturbation degrades a sandbagging policy faster than the underlying capability, so an improvement under noise
might itself be evidence of suppression.

\subsection{Datasets}
\label{app:datasets}
We describe the main datasets used throughout this paper. MMLU \citep{hendrycks2021measuring} tests world knowledge and problem solving across STEM, humanities, and social sciences. WMDP
\citep{li2024wmdpbenchmarkmeasuringreducing} tests hazardous knowledge relevant to biosecurity, chemical weapons, and cybercrime, and is therefore the benchmark most directly tied to the threat model that motivates capability evaluations. ARC-Challenge \citep{clark2018thinksolvedquestionanswering} tests grade-school science reasoning and provides a third benchmark with a different question distribution.

\section{Localization}
\label{app:localization}

Section~\ref{sec:localization} establishes that ablating the heads identified by
our functional graph recovers suppressed capability better than ablating random
heads. It remains unclear, however, whether this still holds once ablation is
combined with other elicitation methods. Prefill and few-shot prompting already
change how the model processes its input, so after combining with them, ablating
any heads might help, or the choice of heads might no longer matter.

To test this, we repeat the inference-level compositions of
Table~\ref{tab:prompted_sandbagging}, replacing our selected heads with the same
number of randomly chosen heads, and average the results over five independent
random draws. If our selection mattered only for ablation alone, random heads
should perform comparably once combined with prefill or few-shot prompting.

Table~\ref{tab:localization_random_controls} shows that they do not. Random
heads give inconsistent gains and often reduce accuracy below the baseline
alone, while our selection improves performance in nearly every case. The
benefit of our method therefore comes from where we intervene, even when it is
combined with existing elicitation methods.

\begin{table}[!htbp]
\centering
\caption{Localization controls under prompted sandbagging. Values are
final-evaluation accuracy (\%). Random-head results report the mean and sample
standard deviation over five random-head draws with the same intervention
budget as ours. Interventions on random heads yield inconsistent gains and in
some cases reduce accuracy. This shows how our method identifies attention
heads that enable targeted intervention.}
\label{tab:localization_random_controls}

\small
\renewcommand{\arraystretch}{1.12}
\newcommand{\locscore}[1]{\ensuremath{#1}}
\newcommand{\locscoresd}[2]{\ensuremath{#1_{\scriptscriptstyle\pm #2}}}
\newcommand{\locbest}[1]{\ensuremath{\boldsymbol{#1}}}

\begin{tabular}{@{}lllll@{}}
\toprule
\textbf{Model}
& \textbf{Method}
& \multicolumn{1}{l}{\textbf{MMLU}}
& \multicolumn{1}{l}{\textbf{WMDP}}
& \multicolumn{1}{l}{\textbf{ARC}}
\\
\midrule

\multirow{6}{*}{\textbf{Mistral-24B}}
& Prefill
& \locscore{30.75} & \locscore{28.25} & \locscore{33.75}
\\
& \textcolor{baselinegray}{Random heads + Prefill}
& \textcolor{baselinegray}{\locscoresd{30.55}{1.73}}
& \textcolor{baselinegray}{\locscoresd{28.05}{0.99}}
& \textcolor{baselinegray}{\locscoresd{34.75}{2.86}}
\\
& \textbf{Ours + Prefill}
& \locbest{36.75} & \locbest{33.75} & \locbest{45.50}
\\
& Five-shot
& \locscore{15.58} & \locscore{19.92} & \locscore{6.42}
\\
& \textcolor{baselinegray}{Random heads + Five-shot}
& \textcolor{baselinegray}{\locscoresd{16.10}{0.64}}
& \textcolor{baselinegray}{\locscoresd{20.45}{0.94}}
& \textcolor{baselinegray}{\locscoresd{7.17}{0.86}}
\\
& \textbf{Ours + Five-shot}
& \locbest{19.42} & \locbest{22.58} & \locbest{9.50}
\\

\midrule

\multirow{6}{*}{\textbf{Qwen3-32B}}
& Prefill
& \locscore{67.75} & \locscore{60.75} & \locscore{89.50}
\\
& \textcolor{baselinegray}{Random heads + Prefill}
& \textcolor{baselinegray}{\locscoresd{68.15}{2.65}}
& \textcolor{baselinegray}{\locscoresd{60.30}{1.81}}
& \textcolor{baselinegray}{\locscoresd{88.65}{1.36}}
\\
& \textbf{Ours + Prefill}
& \locbest{68.00} & \locbest{62.00} & \locbest{91.25}
\\
& Five-shot
& \locscore{73.00} & \locscore{63.83} & \locscore{87.08}
\\
& \textcolor{baselinegray}{Random heads + Five-shot}
& \textcolor{baselinegray}{\locscoresd{73.98}{1.50}}
& \textcolor{baselinegray}{\locscoresd{65.20}{1.69}}
& \textcolor{baselinegray}{\locscoresd{88.55}{3.07}}
\\
& \textbf{Ours + Five-shot}
& \locbest{76.50} & \locbest{68.33} & \locbest{93.58}
\\

\midrule

\multirow{6}{*}{\textbf{Llama-3.3-70B}}
& Prefill
& \locscore{27.75} & \locscore{17.00} & \locscore{10.75}
\\
& \textcolor{baselinegray}{Random heads + Prefill}
& \textcolor{baselinegray}{\locscoresd{25.45}{3.06}}
& \textcolor{baselinegray}{\locscoresd{16.35}{1.38}}
& \textcolor{baselinegray}{\locscoresd{10.25}{0.94}}
\\
& \textbf{Ours + Prefill}
& \locbest{31.75} & \locbest{20.50} & \locbest{21.50}
\\
& Five-shot
& \locscore{20.00} & \locscore{22.42} & \locscore{12.58}
\\
& \textcolor{baselinegray}{Random heads + Five-shot}
& \textcolor{baselinegray}{\locscoresd{24.70}{4.61}}
& \textcolor{baselinegray}{\locscoresd{25.53}{3.42}}
& \textcolor{baselinegray}{\locscoresd{17.92}{5.21}}
\\
& \textbf{Ours + Five-shot}
& \locbest{47.08} & \locbest{39.00} & \locbest{46.92}
\\

\bottomrule
\end{tabular}
\end{table}

\paragraph{Zero Ablation vs.\ GraphLearned: Random and Graph-Selected Heads.}
We compare zero ablation and GraphLearned using random and graph-selected heads under prompted sandbagging on Mistral-24B and Qwen3-32B. We find that graph-selected head ablation consistently outperforms random ablation across all datasets for both zero ablation and GraphLearned experiments. Adding learnable parameters further improves performance. These results further support the benefits of graph-based head selection.

\begin{table}[!htb]
\centering
\caption{Zero ablation versus GraphLearned under random and graph-based head selection. Values are final-evaluation accuracy (\%). Error terms denote sample SD across five head sets for random ablation, five head-set means for random GraphLearned, and five training trials for graph-selected GraphLearned. Graph-selected ablation uses one fixed head set.}
\label{tab:zero_vs_graphlearned}
\small
\renewcommand{\arraystretch}{1.12}
\newcommand{\zgscore}[1]{\ensuremath{#1}}
\newcommand{\zgbest}[1]{\ensuremath{\boldsymbol{#1}}}
\newcommand{\zgscoresd}[2]{\ensuremath{#1_{\scriptscriptstyle\pm #2}}}
\newcommand{\zgbestsd}[2]{\ensuremath{\boldsymbol{#1_{\scriptscriptstyle\pm #2}}}}
\begin{tabular}{@{}lllll@{}}
\toprule
\textbf{Model} & \textbf{Method} & \multicolumn{1}{l}{\textbf{MMLU}} & \multicolumn{1}{l}{\textbf{WMDP}} & \multicolumn{1}{l}{\textbf{ARC}} \\
\midrule
\multirow{4}{*}{\textbf{Mistral-24B}}
& Zero ablation (random) & \zgscoresd{22.40}{2.46} & \zgscoresd{22.15}{2.39} & \zgscoresd{14.05}{3.26} \\
& \textbf{Zero ablation (graph)} & \zgbest{53.00} & \zgbest{45.50} & \zgbest{60.50} \\
& GraphLearned (random heads) & \zgscoresd{55.11}{5.76} & \zgscoresd{49.97}{2.60} & \zgscoresd{64.13}{8.73} \\
& \textbf{GraphLearned (graph-selected heads)} & \zgbestsd{71.35}{2.80} & \zgbestsd{60.70}{0.82} & \zgbestsd{89.85}{3.21} \\
\midrule
\multirow{4}{*}{\textbf{Qwen3-32B}}
& Zero ablation (random) & \zgscoresd{27.05}{4.19} & \zgscoresd{28.10}{6.29} & \zgscoresd{16.90}{6.37} \\
& \textbf{Zero ablation (graph)} & \zgbest{64.75} & \zgbest{61.25} & \zgbest{81.50} \\
& GraphLearned (random heads) & \zgscoresd{49.65}{4.33} & \zgscoresd{47.62}{3.89} & \zgscoresd{52.05}{9.98} \\
& \textbf{GraphLearned (graph-selected heads)} & \zgbestsd{70.25}{1.33} & \zgbestsd{63.60}{1.10} & \zgbestsd{89.10}{3.18} \\
\bottomrule
\end{tabular}
\end{table}

\FloatBarrier
\section{Comparison across demonstrations}
\label{app:comparison_across_demonstrations}

The weight-level results in Table~\ref{tab:prompted_sandbagging} are obtained
with four demonstrations. Since full fine-tuning updates far more parameters, it
may benefit more from additional data. We therefore examine whether our
advantage persists as the demonstration budget increases, comparing
\textsc{Ours (GraphLearned)} with full fine-tuning at four, eight, and sixteen
demonstrations over five independent trials.

\begin{table}[!htb]
\centering
\caption{Comparison across demonstration budgets. Values are final-evaluation
accuracy (\%), reported as mean and sample standard deviation over five
independent trials. Bold marks the higher mean within each matched model,
demonstration budget ($m$), and benchmark comparison. Our method outperforms
Full-SFT in nearly all scenarios across both lower and higher demonstration
budgets and has lower variance in every comparison.}
\label{tab:comparison_across_demonstrations}

\small
\renewcommand{\arraystretch}{1.12}
\newcommand{\demscore}[2]{\ensuremath{#1_{\scriptscriptstyle\pm #2}}}
\newcommand{\dembest}[2]{\ensuremath{\boldsymbol{#1_{\scriptscriptstyle\pm #2}}}}

\begin{tabular}{@{}lll@{\hspace{15pt}}lll@{}}
\toprule
\textbf{Model}
& $\boldsymbol{m}$
& \textbf{Method}
& \multicolumn{1}{l}{\textbf{MMLU}}
& \multicolumn{1}{l}{\textbf{WMDP}}
& \multicolumn{1}{l}{\textbf{ARC}}
\\
\midrule

\multirow{6}{*}{\textbf{Mistral-24B}}
& \multirow{2}{*}{$4$}
& Full-SFT
& \demscore{68.05}{9.43}
& \demscore{58.60}{7.50}
& \demscore{86.10}{9.86}
\\
& & GraphLearned
& \dembest{71.35}{2.80}
& \dembest{60.70}{0.82}
& \dembest{89.85}{3.21}
\\
\addlinespace[2pt]
& \multirow{2}{*}{$8$}
& Full-SFT
& \demscore{69.85}{1.83}
& \demscore{59.10}{2.56}
& \demscore{89.35}{1.76}
\\
& & GraphLearned
& \dembest{72.20}{0.74}
& \dembest{61.25}{0.98}
& \dembest{91.30}{0.48}
\\
\addlinespace[2pt]
& \multirow{2}{*}{$16$}
& Full-SFT
& \demscore{67.90}{3.34}
& \demscore{60.30}{4.08}
& \demscore{89.45}{2.00}
\\
& & GraphLearned
& \dembest{72.70}{1.05}
& \dembest{62.00}{0.68}
& \dembest{91.45}{0.21}
\\

\midrule

\multirow{6}{*}{\textbf{Qwen3-32B}}
& \multirow{2}{*}{$4$}
& Full-SFT
& \demscore{63.55}{19.33}
& \demscore{54.80}{14.50}
& \demscore{76.95}{32.40}
\\
& & GraphLearned
& \dembest{70.25}{1.33}
& \dembest{63.60}{1.10}
& \dembest{89.10}{3.18}
\\
\addlinespace[2pt]
& \multirow{2}{*}{$8$}
& Full-SFT
& \demscore{66.00}{13.09}
& \demscore{59.50}{6.04}
& \demscore{79.35}{24.97}
\\
& & GraphLearned
& \dembest{69.75}{2.68}
& \dembest{64.05}{0.93}
& \dembest{88.90}{4.26}
\\
\addlinespace[2pt]
& \multirow{2}{*}{$16$}
& Full-SFT
& \dembest{73.75}{2.01}
& \demscore{64.05}{0.86}
& \demscore{91.80}{0.97}
\\
& & GraphLearned
& \demscore{71.45}{0.78}
& \dembest{64.90}{0.60}
& \dembest{92.00}{0.64}
\\

\bottomrule
\end{tabular}
\end{table}

As shown in Table~\ref{tab:comparison_across_demonstrations}, our method achieves
higher accuracy in nearly all comparisons and lower variance in every one. The
variance of full fine-tuning decreases with more demonstrations, whereas our
method is already stable at the smallest budget. On Mistral-24B, our method with
four demonstrations exceeds full fine-tuning with sixteen demonstrations. These results
indicate that our method recovers the suppressed capability reliably with fewer
demonstrations and remains competitive as more become available.

\FloatBarrier
\section{Comparison with Activation-Based Head Selection}
\label{app:activation_comparison}

We compare our graph-based head selection with an alternative approach that selects heads based on their normalized activation changes between honest and sandbagging conditions. We use equal head budgets and apply identical zero-ablation interventions. We evaluate both approaches across nine sandbagging model settings, including five additional sandbagging-trained models from \citet{greenblatt2024stress} and \citet{taylor2025auditing}.

As shown in the table below, graph-based selection achieves higher capability elicitation accuracy across all nine settings, with improvements of up to 50 percentage points on Willow and 32.43 percentage points on circuit-broken Mistral Instruct. These results demonstrate that graph-based selection identifies more effective heads for capability elicitation than selection based solely on activation changes.

\begin{table}[!htb]
\centering
\caption{Capability elicitation accuracy (\%) using activation-based versus graph-based head selection under identical head-ablation budgets.}
\label{tab:activation_vs_graph}
\small
\renewcommand{\arraystretch}{1.12}
\begin{tabular}{@{}lrrr@{}}
\toprule
\textbf{Sandbagging Model} & \textbf{Activation (\%)} & \textbf{Graph (\%)} & \textbf{Graph $-$ Activation (pp)} \\
\midrule
Redwood DeepSeek-Math-7B/Pythia-1B & 7.29 & \textbf{10.94} & +3.65 \\
Gemma-2-9B Code & 25.00 & \textbf{27.34} & +2.34 \\
Willow & 36.72 & \textbf{86.72} & +50.00 \\
Tarun & 57.30 & \textbf{59.12} & +1.82 \\
Cedar & 73.47 & \textbf{80.61} & +7.14 \\
Password-locked Mistral Base & 56.49 & \textbf{57.57} & +1.08 \\
Password-locked Mistral Instruct & 50.00 & \textbf{51.35} & +1.35 \\
Circuit-broken Mistral Base & 59.46 & \textbf{61.08} & +1.62 \\
Circuit-broken Mistral Instruct & 4.32 & \textbf{36.76} & +32.43 \\
\bottomrule
\end{tabular}
\end{table}

\FloatBarrier
\begingroup
\color{revisionblue}
\section{Understanding Graph-Localized Ablation}
\label{app:understanding_ablation}

We investigate why ablating graph-selected heads recovers suppressed capabilities while largely preserving honest accuracy, using prompted sandbagging on Mistral-24B and Qwen3-32B. One possible explanation is that these heads might be contributing little to normal computation and removing them has minimal effect on honest performance.
To test this, we measure each head's activation magnitude before the output projection, $\|z_h\|_2$, and its contribution magnitude after the projection, $\|W_O^{(h)}z_h\|_2$, at the final prompt token under both honest and sandbagging conditions. We average these magnitudes across evaluation questions for each head.
For comparison, we construct 20 random head sets, each containing exactly the same number of heads per layer as our graph-selected set. We measure their magnitudes on the same questions and compute the mean magnitude across each set. For each condition and magnitude measure, we report the ratio between the mean magnitude of graph-selected heads and the average magnitude across the 20 random sets.
As shown in Table~\ref{tab:head_magnitude}, graph-selected heads have magnitudes generally comparable to randomly selected heads, and in some settings are even larger. These results shows that our selected heads are not simply low-magnitude components.

\begin{table}[!htbp]
\centering
\color{revisionblue}
\caption{Mean activation magnitudes of graph-selected heads relative to
layer-matched random heads, before and after the output projection, under
honest and sandbagging conditions.}
\label{tab:head_magnitude}
\begingroup
\small
\setlength{\tabcolsep}{7pt}
\begin{tabular}{@{}lcccc@{}}
\toprule
& \multicolumn{2}{c}{\textbf{Honest}}
& \multicolumn{2}{c}{\textbf{Sandbagging}} \\
\cmidrule(lr){2-3}
\cmidrule(lr){4-5}
\textbf{Model}
& \textbf{Pre-$W_O$}
& \textbf{Post-$W_O$}
& \textbf{Pre-$W_O$}
& \textbf{Post-$W_O$} \\
\midrule
Mistral-24B & $1.11\times$ & $1.09\times$ & $1.17\times$ & $1.15\times$ \\
Qwen3-32B   & $1.22\times$ & $1.20\times$ & $1.14\times$ & $1.12\times$ \\
\bottomrule
\end{tabular}
\endgroup
\end{table}

We further test whether head magnitude alone can explain capability elicitation. We construct 20 random head sets, each containing the same number of heads per layer as our graph-selected set. Within each layer, we select heads with projected-output magnitudes close to those of the corresponding graph-selected heads, measured under the honest condition.
We then ablate these heads using the same intervention and evaluate their ability to recover suppressed performance. 
As shown in Figure~\ref{fig:magnitude_matched_ablation}, magnitude-matched random ablation fails to reproduce the recovery achieved by graph selection. This suggests that head magnitude alone cannot explain the effectiveness of elicitation.

\begin{figure}[!hb]
\centering
\color{revisionblue}
\includegraphics[width=0.7\linewidth]{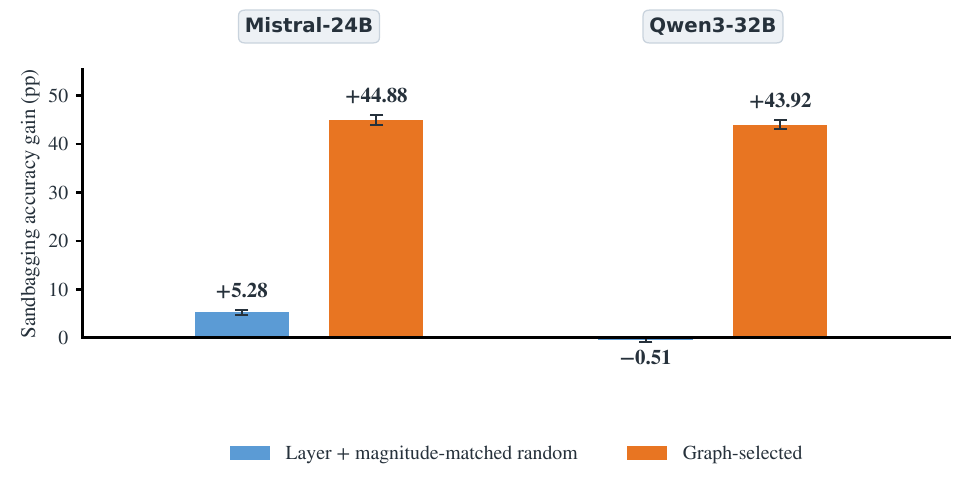}

\caption{MMLU sandbagging accuracy gains from ablating graph-selected heads
versus the mean of layer- and magnitude-matched random head sets. Error bars
indicate 95\% paired-question bootstrap confidence intervals with head sets
held fixed. Graph-selected ablation achieves substantially greater capability
ellicitation than matched random ablation.}

\label{fig:magnitude_matched_ablation}
\end{figure}

\subsection{Representational Changes Under Ablation}
\label{app:representation_ablation}

\begin{figure}[!hb]
\centering
\color{revisionblue}
\includegraphics[width=0.9\linewidth]{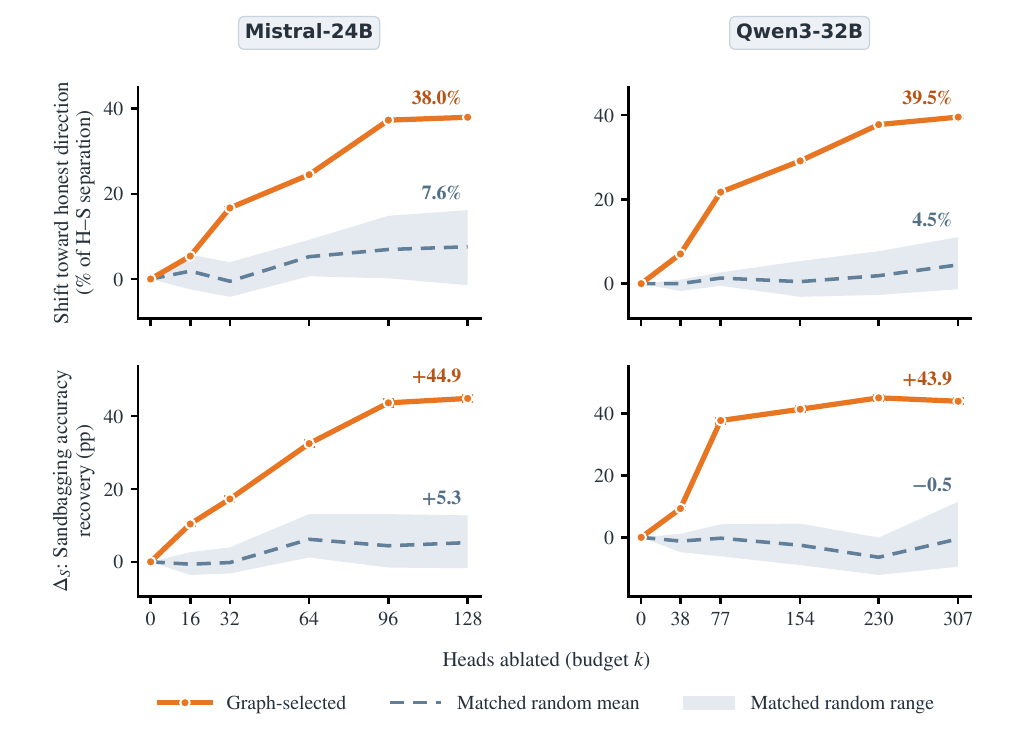}
\caption{Effects of increasing head ablation on MMLU. Top: representational shift toward the honest condition along the honest--sandbagging contrast direction (\% of separation). 
Bottom: sandbagging accuracy gain (percentage points). Graph-selected ablation produces larger representational shifts and greater capability recovery than layer- and magnitude-matched random ablation.}
\label{fig:representation_dose}
\end{figure}

Motivated by our finding that head magnitude alone does not explain the capability recovery achieved by graph-localized ablation, we further analyse how the intervention affects the model's internal representations. 
Specifically, we examine whether ablating graph-selected heads shifts the model's representations toward those observed under honest prompting, and how these shifts relate to the recovery of suppressed performance.

We begin by identifying a direction in the residual stream that distinguishes honest from sandbagging behavior. 
For each question $x$, we run the model under both honest ($H$) and sandbagging ($S$) prompting and record the residual-stream representation at the final prompt token after transformer block $\ell$, denoted by $r_\ell^H(x)$ and $r_\ell^S(x)$, respectively. 
Using a separate set of discovery questions, we compute the average difference between these representations:
\begin{equation}
d_\ell = \mathbb{E}_{x}\left[r_\ell^H(x)-r_\ell^S(x)\right].
\end{equation}
This defines a fixed honest--sandbagging contrast direction at each measured block. It provides a reference axis along which we can characterize representational changes, independently of the ablation outcomes evaluated below.

Next, on held-out evaluation questions, we repeat the sandbagging run while ablating a selected set of attention heads $A$ at the final prompt token. We denote the resulting residual-stream representation by $r_\ell^{S,A}(x)$. Comparing this representation with the original, unablated sandbagging representation allows us to measure the change induced by the intervention. We project this change onto the previously defined contrast direction:
\begin{equation}
s_\ell(x;A) =
\frac{\left(r_\ell^{S,A}(x)-r_\ell^S(x)\right)^\top d_\ell}
{\|d_\ell\|_2^2}.
\end{equation}
A positive value indicates that ablation moves the representation in the direction of the honest condition, while a negative value indicates movement in the opposite direction. The normalization expresses this movement relative to the average honest--sandbagging separation along the contrast direction estimated from the discovery questions. 
For visualization, we express the mean projected shift as a percentage of the corresponding honest--sandbagging separation measured on the held-out questions. This measures movement along a specific direction, rather than overall similarity between the two representations.
To examine how these changes develop as more heads are removed, we progressively increase the number of ablated graph-selected heads. At each ablation budget, we compare the resulting representational shift against random head sets matched in layer distribution and honest-condition projected-output magnitude. We also measure the corresponding change in MMLU accuracy under sandbagging, relative to the unablated model. Figure~\ref{fig:representation_dose} presents both measurements across increasing ablation budgets, allowing us to examine how representational shifts and capability recovery vary together. This analysis characterizes the effects associated with graph-localized ablation without assuming that the measured representational shift is itself the mechanism responsible for recovery.

Overall, we observe that progressively ablating graph-selected heads shifts internal representations increasingly toward the honest condition along the honest--sandbagging contrast direction. These shifts are accompanied by greater recovery in sandbagging accuracy. 
Both effects are largely absent under layer- and magnitude-matched random ablation.

\FloatBarrier
\endgroup

\end{document}